\documentclass[conference]{IEEEtran}
\usepackage{times}
\usepackage{graphicx}
\usepackage{amsmath}
\usepackage{amsfonts}
\usepackage{gensymb}

\usepackage[numbers]{natbib}
\usepackage{multicol}
\usepackage[bookmarks=true]{hyperref}

\begin{document}

\title{CraterGrader: Autonomous Robotic Terrain Manipulation for Lunar Site Preparation and Earthmoving}

\author{\IEEEauthorblockN{Ryan Lee\IEEEauthorrefmark{1}\IEEEauthorrefmark{2}, Benjamin Younes\IEEEauthorrefmark{1}\IEEEauthorrefmark{2}, Alexander Pletta\IEEEauthorrefmark{1}\IEEEauthorrefmark{2}, John Harrington\IEEEauthorrefmark{1}\IEEEauthorrefmark{2}, Russell Q. Wong\IEEEauthorrefmark{1}\IEEEauthorrefmark{2},\\ William ``Red" Whittaker\IEEEauthorrefmark{2}}
\IEEEauthorblockA{\IEEEauthorrefmark{1}Equal Contribution, Listed in Random Order}
\IEEEauthorblockA{\IEEEauthorrefmark{2}Carnegie Mellon University. 
}}%



%

\maketitle

\begin{abstract}

Establishing lunar infrastructure is paramount to long-term habitation on the Moon. To meet the demand for future lunar infrastructure development, we present CraterGrader, a novel system for autonomous robotic earthmoving tasks within lunar constraints. In contrast to the current approaches to construction autonomy, CraterGrader uses online perception for dynamic mapping of deformable terrain, devises an energy-efficient material movement plan using an optimization-based transport planner, precisely localizes without GPS, and uses integrated drive and tool control to manipulate regolith with unknown and non-constant geotechnical parameters. We demonstrate CraterGrader's ability to achieve unprecedented performance in autonomous smoothing and grading within a lunar-like environment, showing that this framework is capable, robust, and a benchmark for future planetary site preparation robotics.
\end{abstract}

\IEEEpeerreviewmaketitle

\section{Introduction}
As humanity sets its sights on the Moon, Mars, and beyond, human presence requires surface infrastructure similar to the buildings and roads relied upon for millennia on Earth \cite{NASADIRECTIVE} \cite{LUSTR}. Robotic systems show promise in constructing surface infrastructure in aerospace applications, including landing pads, roads, structural foundations, trenches, and berms, particularly in locations such as the Moon or Mars. However, the technological demands of operating in the low-mass \cite{doi:10.1177/0278364915615689}, low-energy, and high-stakes environment of space require new approaches to robotic construction. Challenges such as time delays and lack of environmental awareness in teleoperation, can be overcome through increased autonomy \cite{miller2005visual}. Continuous operation time is crucial when vehicle lifetime and operationally-constrained energetics are limited. To this end, autonomy is capable of operating patiently and continuously over a 14-day lunar daylight period with minimal human supervision. Thus, there is a pressing need to develop autonomous robots that can efficiently and effectively perform lunar site preparation.

\begin{figure}[]
    \centering
    \includegraphics[width=0.45\textwidth]{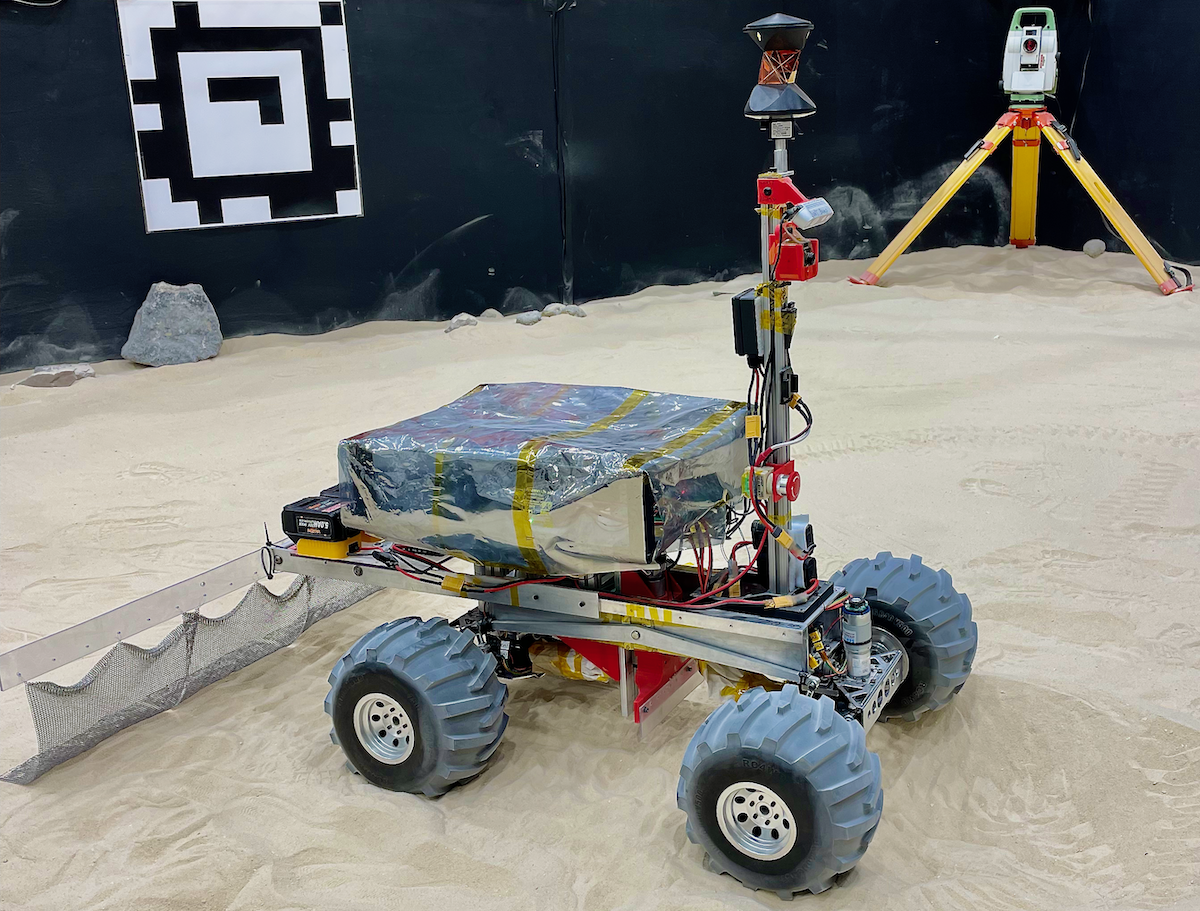}
    \caption{CraterGrader in the MoonYard. Top left: Pseudo-sun fiducial tag used solely for bearing estimation. Center: CraterGrader. Top right: Leica Viva TS16 Total Station.}
    \label{fig:glory}
\end{figure}

We present CraterGrader: a novel approach to autonomous robotics for lunar site preparation (Figure \ref{fig:glory}). The main contribution of this work is a  combination of the following: (1) System demonstration of autonomous lunar grading meeting NASA's LuSTR RFP specifications \cite{LUSTR} with no priors, (2) Novel application of optimal material movement planning for autonomous construction, and (3) GPS-free precision robot localization subject to planetary site preparation constraints. We also discuss the experimental performance and results of the manifested robotic system, emphasizing its potential applications in lunar infrastructure development. Our work demonstrates an approach to delivering a site preparation robot that can autonomously manipulate lunar-like regolith in a worksite to form desired topography.

\section{Related Work}

\subsection{Earth Moving Autonomy}

Site preparation refers to the tasks that produce a baseline worksite for building foundations and superstructures. Flattening and smoothing the terrain to meet a specified planar angle, also called ``grading", is a time-intensive component of construction that requires precision. Although grading is a common task in most construction projects, there is limited research on autonomy in this area, despite significant industry interest \cite{MELENBRINK2020103312}. 

Early work demonstrated an initial framework for intelligent earth working \cite{KIM20031}, focusing on developing a system architecture and hierarchy of the proposed concept. \citet{BONCHIS201111588} developed an autonomous skid-steer loader, supported with a sensor suite, onboard computer, and limited supervised autonomy, highlighting the potential of an automated solution in the task of bulk material handling. Later, \citet{Marshall2008562} and \citet{Schmidt5509546} presented a number of approaches to the dig and dump planning processes, using a 2.5D (height-field) representation of the environment to plan digging. \citet{JUD2021103783} developed a robotic walking excavator, HEAP, retrofitted from a commercial excavator and demonstrated autonomy in excavation tasks, including the formation of free-form embankments. State-of-the-art systems deployed in the construction industry have focused primarily on assistive features \cite{caterpillar} or selected subtasks within the greater construction industry \cite{built}.  

Unlike terrestrial priors, our approach is energy-efficient within the constraints of lunar applications and does not rely on existing infrastructure such as GPS.

\subsection{Lunar Construction}

The field of lunar construction and In Situ Resource Utilization (ISRU) has grown significantly in the 21st century, as the goal of a presence on the Moon has become more in reach \cite{NASADIRECTIVE}. In 2004, \citet{muff2004prototype} demonstrated an initial prototype bucket wheel excavator. This was later expanded upon by the NASA Swamp Works team at Kennedy Space Center in 2010 \cite{schuler2019rassor}, where the RASSOR platform was developed using the principle of a counter-rotating bucket drum to enable strong cutting forces with reduced weight-on-tool. Lunar site preparation was explored in 2009 during the integration of a front-facing lightweight bulldozer blade onto NASA's Chariot platform \cite{mueller2009lightweight}.

Recently, more attention has been given to developing lightweight robotic systems \cite{skonieczny2010parameters}. In 2014, a lightweight robotic excavator was demonstrated to carry over 50\% of its weight \cite{red_lightweight}. In 2018, a teleoperated platform was developed for site preparation that graded and leveled a basalt rock simulated lunar worksite \cite{kelso2016planetary}.  A rare example of autonomy research within lunar excavation focused on an evolutionary learning approach to planning, as well as ground truth localization and mapping information provided from an external source \cite{thangavelautham2017autonomous}.

In 2021, a NASA Lunar Surface Technology Research (LuSTR) request for proposal (RFP) outlined a desire for an autonomous lightweight system that could grade, compact, and remove rocks from a lunar worksite for a lunar landing pad \cite{LUSTR}. The RFP emphasized the lack of research in lunar site preparation autonomy and provided a standard reference for lunar site preparation requirements. These requirements directly inspired this project's goals and system design, providing a framework for effective evaluation.

\subsection{Lunar Localization \& Mapping}

Localization is one of the most difficult challenges within planetary robotics due to the lack of infrastructure, harsh lighting conditions, and unknown environments. Past missions, such as the Mars Exploration Rovers, heavily relied on Visual Odometry \cite{cheng2005visual} and a higher tolerance for error. Ultra-wideband (UWB) beacons are of increasing interest for Time Distance of Arrival positioning in lunar environments \cite{jiang2008lunar}.

Other localization methods include using neural networks for global map correspondence \cite{wu2020absolute} and crater landmark detection \cite{9843714}. However, in the context of site preparation, relying on such landmarks for localization is not feasible as they would be inherently altered during terrain manipulation processes. \citet{molina2012lander} investigated the use of robotic total stations for line-of-sight tracking as part of a lander subsystem. A robotic total station was also used in a lunar prospector platform, Scarab \cite{wettergreen2010design}, resulting in precise localization and robust slip estimation at 1 Hz. Strict vertical height tolerances for mapping and grading necessitate 3D positioning to a precision that is unprecedented in prior planetary robotics applications.

\subsection{Optimal Transport}

Optimal transport refers to the problem of transforming one distribution into another using an \textit{Optimal Transport Plan}, which minimizes the cost of transforming between the two distributions, also known as the Wasserstein distance. Gaspard Monge conceptualized the field of optimal transport in 1781 for transportation theory in the context of soil transport \cite{monge1781memoire}. Optimal transport has practical applications in fields such as networking \cite{burger2021dynamic}, image retrieval \cite{rubner2000earth}, graphics \cite{solomon2014earth}, and computer vision \cite{gottschlich2014separating} \cite{6216380}. An optimal transport plan represents a function map that transforms between the distributions through a minimally exertive process, analogous to mechanical work. \citet{zhai2019path} used optimal transport maps for path planning in unknown environments. The name ``Earth Mover's Distance" (EMD) was coined by Jorge Stolfi in 1994 and is synonymous with the Wasserstein distance \cite{rubner2000earth}. The term ``Earth Mover's Distance" closely matches our work in a most literal sense, in which we introduce a novel application of EMD in the field of robotics and earth moving.

\section{System Architecture}

\subsection{System Objective}\label{sec:system_objective}
\begin{figure}[]
    \medskip
    \centering
    \includegraphics[width=0.45\textwidth]{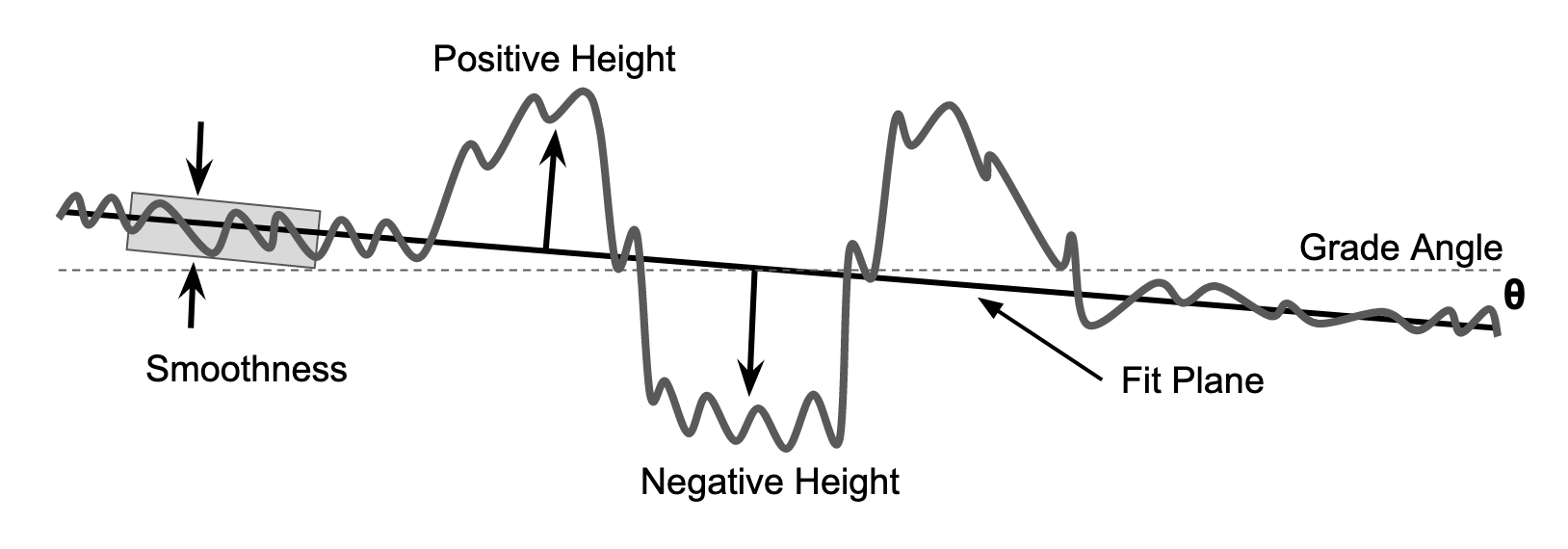}
    \caption{An example crater cross-section, depicting key parameters such as grade, smoothness, and positive/negative height, with respect to a fit ground plane. In particular, grade and smoothness metrics are used to evaluate CraterGrader's performance against the LuSTR RFP baseline \cite{LUSTR}. Grade is defined as the angle between the fit plane and a plane normal to gravity. Smoothness is calculated by constructing a distribution of positive and negative terrain heights relative to the fit plane and computing its standard deviation. The roughness of the topography and grade angle are not to scale and are exaggerated for effect.}
    \label{fig:crater-profile}
\end{figure}

The main objective of CraterGrader is to change the local grade, absolute height, and surface smoothness for a given \textit{worksite} of previously unseen lunar-like topography into the desired flat terrain topography. Here, the term \textit{grade} refers to the slope, or gradient, of an assumed continuous surface topography. Grade, measured in degrees, is defined as the angle between a fit plane to the terrain and a level plane normal to gravity. Smoothness, measured in cm, is computed as the standard deviation of the distribution of positive and negative terrain heights relative to the fit plane. Visualizations of these metrics against a crater cross-section are shown in Figure \ref{fig:crater-profile}. To measure the performance of the system, we derive post-operation grade, with respect to the ground plane, and smoothness requirements of $\pm1\degree$  and 1 cm respectively directly from the LuSTR RFP as a baseline \cite{LUSTR}. We define an additional metric, area out-of-spec (OOS) that measures the worksite area outside of the aforementioned grade and smoothness requirements, reported as a percent reduction after autonomous operation. 

The \textit{worksite} is defined as a rectangular area where work is to be completed. To simplify practical operations, we assume that the worksite has obstacle-free, homogeneous terrain and no other agents or gravity offloading. The parameterization of craters in the worksite assumes a depth-to-diameter ratio of 0.2 m and a crater diameter of 1.0 m, derived from the LuSTR RFP \cite{LUSTR}, and additionally assumes that the volume of material removed from the crater is equal to the volume contained in the crater rim.

The concept of operations is as follows: CraterGrader is placed in the worksite and is given a high-level command to begin autonomous operation. This comprises an initial exploration phase, where no mapping priors are used, followed by a transport phase during which bulk material is moved while grading and smoothing, described in detail in Section \ref{sec:autonomy}. The robot's autonomous operation ceases once all the transport objectives are accomplished. Additionally, the system generates a human-interpretable topographical map of the current worksite, which can be utilized for autonomous planning and also update remote human operators on grading performance and progress.

\subsection{Robot Platform}
 
The robot mobility platform, illustrated in Figure \ref{fig:glory}, has a 0.5 m wheelbase and 0.43 m center-line track width, which fall within a sub-25 kg total mass. The chassis is made of aluminum extrusion and serves as the backbone for the roll-averaging rocker suspension. The wheels consist of a polymer exterior and foam core, but can be swapped with metallic wheels for flight applications. The wheels are driven through the front and rear differentials, with independent steering of the front and rear axles in a double Ackermann configuration. On flat ground, the robot produced 138 N of drawbar pull, as measured by a force transducer. The robot is powered by a rear-mounted battery system, featuring a 20 V 15.0 Ah Li-ion battery with regulation and fusing, which is sufficient for the 100 Watt maximum load on the system. At the front of the robot, a vertical mast provides mounting for various perceptual sensors, and its highest point is approximately 0.7 m from the ground.

The remaining electronics are housed in a dust-protected compartment on the top of the chassis. This compartment includes an NVIDIA Jetson AGX Xavier Development Kit as the main compute system for running the autonomy stack and integrating all peripheral sensing components. An Arduino Due board serves as a lower-level interface for the motor controllers and communicates with the Xavier over a serial connection. All compute systems utilize ROS 2 Galactic middleware to facilitate inter-process communication, with micro-ROS bridging the Xavier and Arduino Due boards.

\subsection{Grading Blade}

\begin{figure}[]
    \medskip
    \centering
    \includegraphics[width=0.45\textwidth]{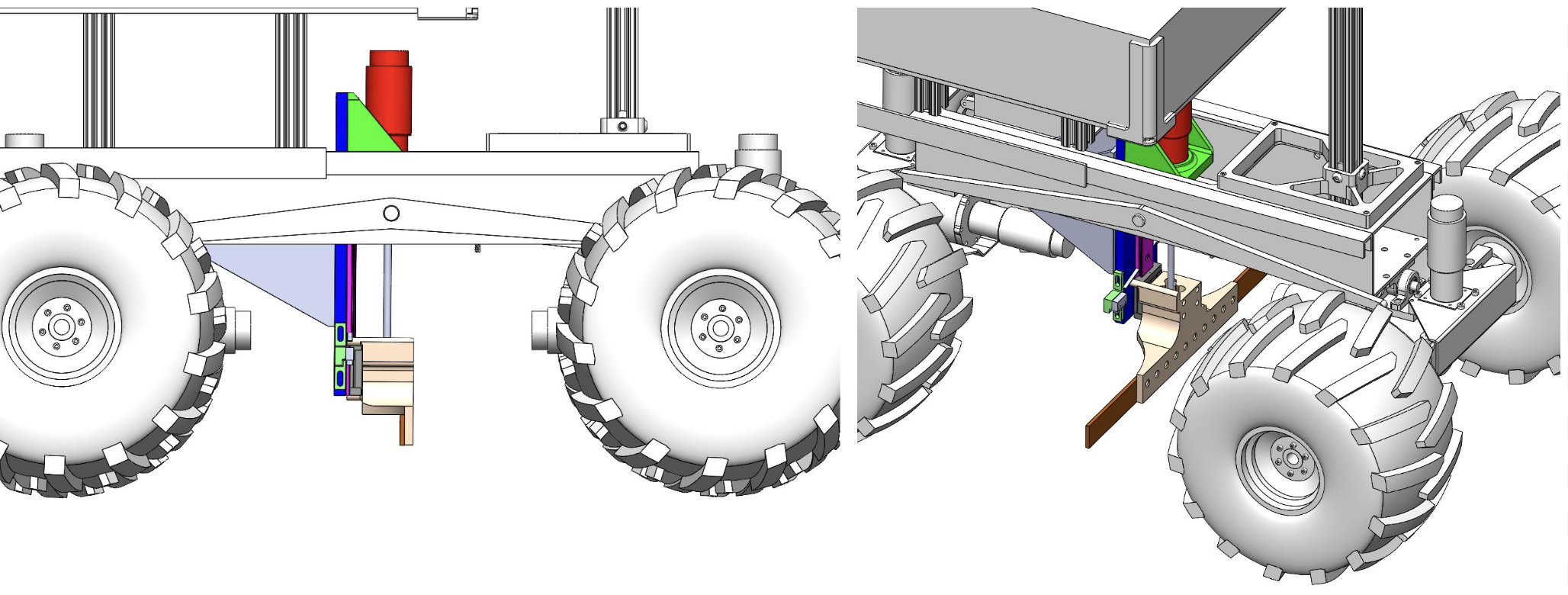}
    \caption{The CraterGrader belly-mounted grading blade assembly. The actuator shown in red is kept inside the robot chassis perimeter to be protected from dust and rock strikes. The blade translates vertically and is rigidly constrained to a linear rail bearing.}
    \label{fig:tool}
\end{figure}

The bulk of CraterGrader's material transport is performed by the grading blade, shown in Figure \ref{fig:tool}. The grading blade has dimensions of 250 mm in length, 50 mm in height, and 3 mm in thickness. The flat blade takes inspiration from terrestrial motor graders and is center-mounted for advantageous drawbar pull, a precious resource in diminished gravity as well as reduced tool gouging on pitched topography. The blade is actuated vertically and can make cuts in the terrain at specified design heights for surface grading and filling craters. This vertical degree of freedom with fast response time enables the grading blade to react dynamically to the terrain as a function of the measured slip or desired cut, and furthermore provides a fallback mechanism for escaping high-centered scenarios. Furthermore, the yaw and pitch angles of the blade can be adjusted statically to experiment with further configurations. In practice, however, maintaining a default blade orientation parallel to the vertical plane yielded the best results, enabling a configuration close to under-vehicle dozing. The grading blade height is relatively shallow to reduce the likelihood of slippage and embedding, allowing excess material to flow over the blade, to be transported in later cuts.

\subsection{Drag Mat}
CraterGrader features a rear-mounted chain mail drag mat, as seen in Figure \ref{fig:glory}. This drag mat effectively smooths high-frequency surface variations, utilizing a similar technique employed in sporting events to smooth terrain, such as a baseball park infield. Notably, the 316L stainless steel drag mat is essential for smoothing out impressions and tire tracks left by the robot during its traversal of the worksite. Although not applicable to all site preparation tasks, our experiments have shown that this drag mat is effective in smoothing out the terrain. Moreover, it is fully passive, simple, and flight-facing with the use of stainless steel.

\subsection{External Infrastructure}\label{sec:external-infra}
Testing for lunar site preparation was conducted in a 56 m$^2$ sandbox referred to as the \textit{MoonYard} (Figure \ref{fig:glory}). The MoonYard features a local network for communication between the robot, the robotic total station, and operators. The regolith simulant used during sandbox testing was fine-grained washed quartz sand with an approximate angle of repose of $31\degree$ and bulk density of 1603 $\frac{kg}{m^3}$. These capabilities facilitate thorough testing and analysis of lunar excavation techniques in a simulated lunar environment.

To provide externally referenced localization estimates, the worksite utilizes a robotic total station and fiducial tags that surround the site, which act as targets for a pseudo-sun tracker. Their usage is described further in Section \ref{sec:sensing}. An NVIDIA TX2 compute system interfaces between the robotic total station and the robot in the worksite, communicating through the local area network.

\section{Autonomy}\label{sec:autonomy}
\subsection{Sensing}\label{sec:sensing}
CraterGrader's sensing suite is designed to operate within the constraints of a lunar environment. An externally-positioned robotic total station tracks a retroreflective prism mounted to the sensing mast of the robot to provide an accurate absolute 3D position estimate with millimeter precision. We present a novel application of the robotic total station to lunar construction, inspired by research application in terrestrial site preparation \cite{garlinge20203d}. This approach is motivated by the absence of GPS and a desire for site-wide consistency. The total station used is a Leica Viva TS16 robotic total station, and is modified to stream measurements at a rate of 7-10 Hz. A flight-forward approach could include a deployable external sensor payload or an additional mobile base for total station setup. A VectorNav VN-100 AHRS (Attitude and Heading Reference System) measures linear acceleration and angular velocity to generate roll and pitch angle estimates. To obtain absolute bearing measurements, sun-tracking sensors are typically used in space applications \cite{sun_sensor}. CraterGrader utilizes a pseudo-sun sensor to mimic this modality in an indoor testing environment by means of a fiducial-tracking fisheye lens camera. An Intel RealSense D435i camera serves as the primary perception sensor, leveraging passive stereo reconstruction to generate point cloud data for terrain analysis. Active stereo through IR illumination was avoided due to infrared scattering expected while operating in dusty lunar terrain. Although lunar dust could degrade depth data in a passive stereo camera, the impact observed during testing was minimal and did not require further mitigation measures. While not space-rated, the RealSense reflects a feasible sensor modality for lunar perception as stereo vision operating in the visual light spectrum. In this mode, the RealSense requires external illumination, and our testing assumes a bright, overhead-lit environment. Lunar deployment may necessitate additional considerations, such as external site lighting or self-illumination on the rover. Explicit crater detection is not applied, allowing for a generic representation of the lunar terrain.

\subsection{Localization}
Operating in a lunar-like environment presents unique challenges for localization in that: (1) the environments are near-featureless, and (2) any prevalent features such as craters are to be destroyed during operation (with the end goal being a featureless, planar surface). Consequently, robust localization that does not rely on traditional SLAM or feature-based methods is required. The 3D pose is estimated by fusing \textit{feature-agnostic} sensor measurements, including live robotic total station positioning, AHRS, pseudo-sun tracker, and onboard encoder telemetry data. Two Extended Kalman Filters (EKFs) are run in parallel to compute both an accurate global position in the \textit{map} frame as well as a smooth, continuous velocity in the \textit{odometry} frame, implemented via the ROS 2 robot\_localization package \cite{robot_localization}. Robot slip is estimated by comparing odometry-measured velocity relative to the absolute position as measured by the total station.

\subsection{Perception \& Mapping}

\begin{figure}[]
    \medskip
    \centering
    \includegraphics[width=0.45\textwidth]{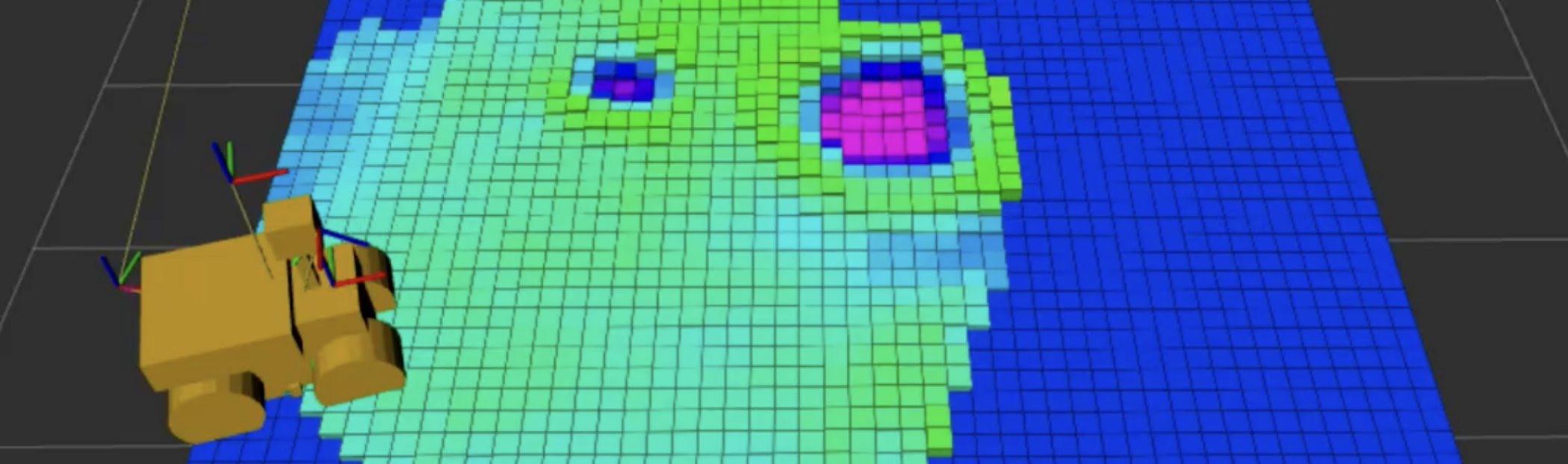}
    \caption{The worksite map is updated as new perception data are acquired. The grid cell color is overloaded in this figure to show both the height of the mapped grid cells and whether a cell has been observed. The dark blue shows map grid cells that have not yet been observed by the robot. Two observed craters are shown alongside the 3D robot model.}
    \label{fig:mappin}
\end{figure}

Point cloud information from stereo reconstruction and state estimates from localization are utilized to generate a 2.5D discretized map of the worksite terrain through a filter-based approach. The system transforms reconstructed 3D points into the map frame and bins them into map grid cells, where each map cell is modeled as a one-dimensional Kalman Filter, akin to \citet{cremean2005uncertainty}. This binned 2.5D map structure provides a lightweight representation of terrain topography which, compared to higher-density point cloud registration methods, is more robust to operating in an environment with sparse features. The confidence estimate of newly seen points is based on localization confidence and measurement distance from the stereo pair. After a cell has been observed a number of times, the robot is increasingly certain about the height in the grid cell until the agent disturbs it with its blade or tires. We model this as noise injection into the grid cell filters after the robot has driven over or manipulated the terrain of those cells. The resulting 2.5D grid map, seen in Figure \ref{fig:mappin}, represents the expected heights at grid cell locations as seen by the robot.

\subsection{Planning}
The planner generates high-level task plans and calculates trajectories for tool operation and the mobility platform to manipulate the terrain based on the robot's state and environment. At the core of the planner is the Behavior Executive, which manages system actions by interfacing with various planners and other software modules. The Behavior Executive also monitors the trajectory-following process and system health, as well as handles faults. The executive is implemented as a hierarchical finite state machine with two levels to switch between operating mode and individual action states, providing a structured approach to the planning and execution of the system's tasks.

The high-level planning strategy involves an initial traversal of the worksite to build up a map of terrain topography prior to earthmoving, followed by an optimal transport plan using the terrain map to perform earthmoving. These steps are handled by the Exploration Planner and Transport Planner respectively, as detailed below.

\subsubsection{Exploration Planner}

Prior to grading, a full map of initial terrain topography is necessary to perform optimal material transport planning as described in Section \ref{sec:transport-planner}. This is achieved by the Exploration Planner, which generates the plan to navigate a worksite of unknown terrain and ultimately build a terrain topography map. First, an operator provides the Exploration Planner with knowledge of the rectangular worksite, including dimensions (length and width) and location of the worksite relative to the robotic total station. The Exploration Planner then generates a set of geometrically-defined waypoints to first navigate the worksite boundaries and then cover the center, to ultimately achieve full map coverage. This strategy proved to be more feasible for the robot's kinematics and large turning radius, compared to other approaches such as a naive raster pattern.

\subsubsection{Transport Planner}\label{sec:transport-planner}
The Transport Planner calculates which \textit{highs} of the worksite should be pushed to which \textit{lows} to convert an arbitrary starting worksite topography to an arbitrary final design topography. Full details on this problem formulation are provided in Section \ref{sec:transport_details}. 

\begin{figure}[]
    \medskip
    \centering
    \includegraphics[width=0.45\textwidth]{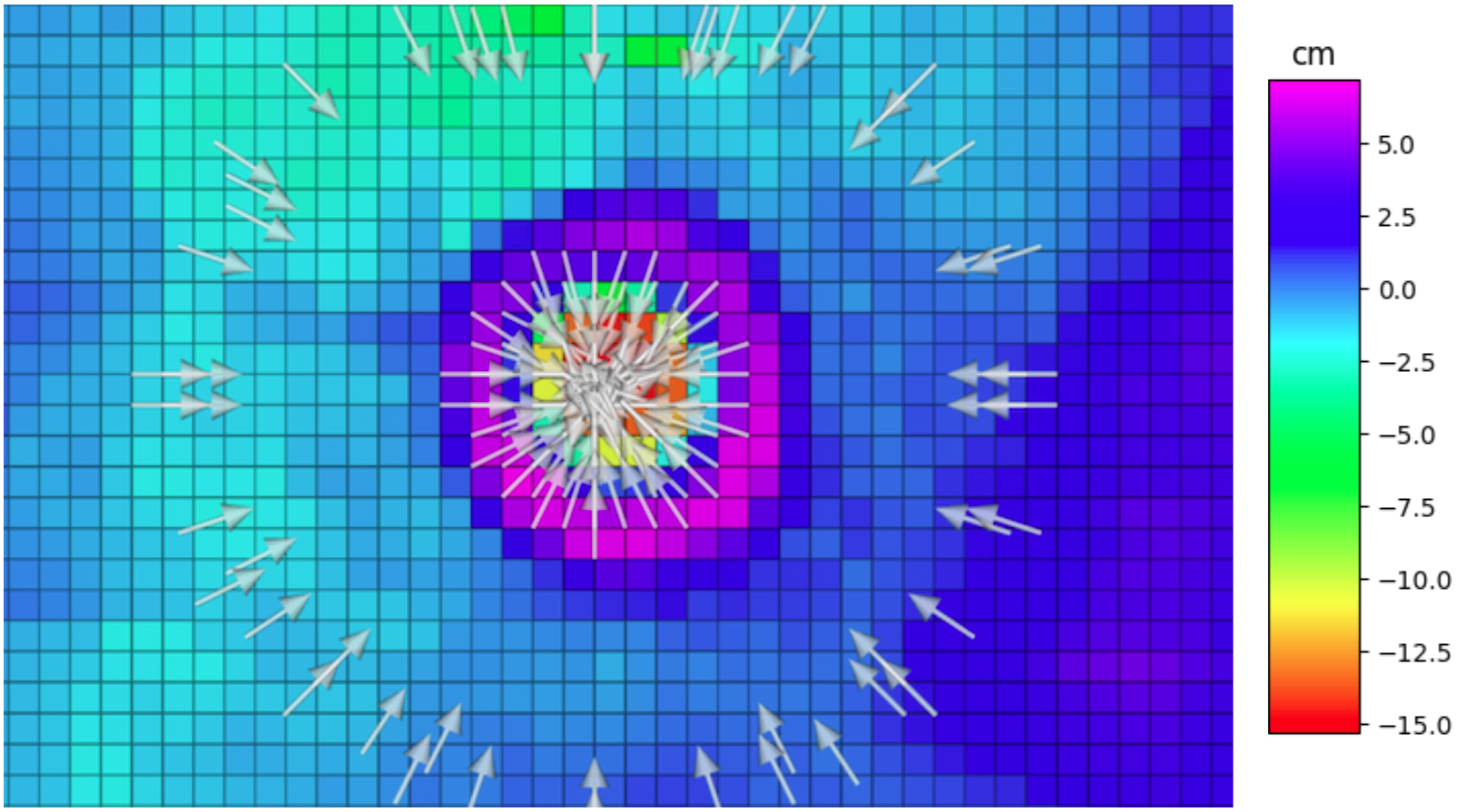}
    \caption{The Transport Planner's optimal online material movement solution for how to fill a real crater. In this diagram, color specifies the height of the grid cell measured from the site-wide fit plane. Red is low at -15.3 cm and purple is high at +7.2 cm. Each white arrow is an element of a \textit{transport triplet}, specifying how the robot should approach the crater and manipulate terrain.}
    \label{fig:transport_with_colorbar}
\end{figure}

This approach was selected instead of motion parameterized by crater geometry to robustly handle malformed craters (as created during operation). Additionally, this approach allows for re-planning for self-corrections, and seamless extension to generalized terraforming. Within the context of grading craters, a design topography was created as a flat plane so the result was a \textit{transport plan} $\Pi$ to pair waypoints along crater rims (\textit{highs}, referred to as \textit{sources} as they contain more material than desired) to waypoints along crater floors (\textit{lows}, referred to as \textit{sinks} as they contain less material than desired), with corresponding material transport volumes. An example transport plan for a real single crater is depicted in Figure \ref{fig:transport_with_colorbar}.


The planner uses linear programming (LP) to solve the problem of minimal-energy terrain manipulation. The formulation is based on an earth mover's distance (EMD) optimization problem, which determines volumes to move between source and sink node pairs, minimizing the total sum of volume moved multiplied by the distances traveled; analogous to mechanical work ignoring transport rate.



The planar coordinates for the pairs of source and sink nodes computed in the transport plan are then used as trajectory waypoints, with the heading aligned to the minimum distance vector between the two waypoints. A third waypoint with the same heading is offset to the rear of the source node to help simplify motion planning, guiding the system to drive in and out of the crater when moving between source and sink pairs. These three waypoints are combined into a \textit{transport triplet} ordered with the offset waypoint first, then the source node, and then the sink node.


After solving the optimization problem online, the Behavior Executive then radially orders the transport triplets and feeds the waypoints in each triplet incrementally to the Kinematic Planner (Section \ref{sec:kinematic-planner}) to calculate the following trajectories.

\subsubsection{Kinematic Planner}\label{sec:kinematic-planner}

\begin{figure}[]
    \medskip
    \centering
    \includegraphics[width=0.45\textwidth]{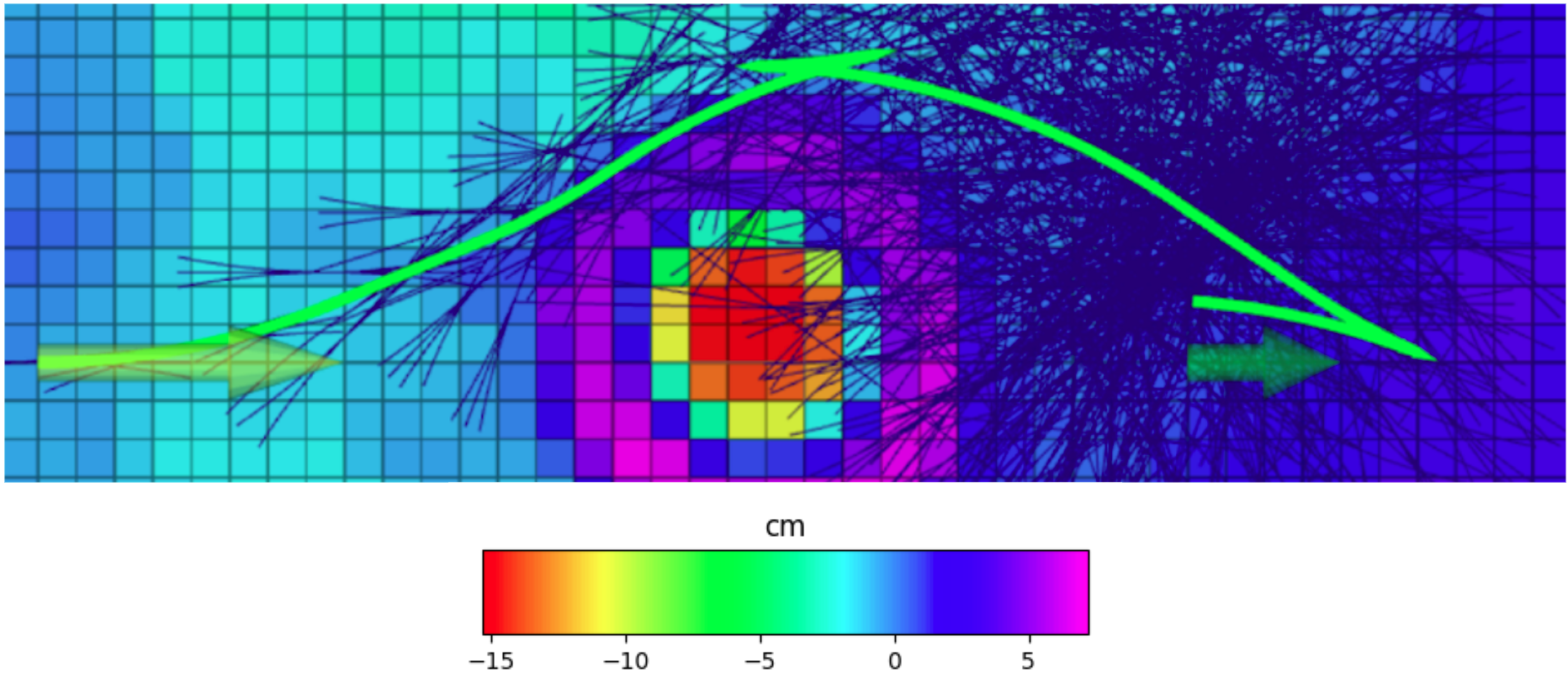}
    \caption{The Kinematic Planner incorporates topography cost to avoid the crater with kinematically feasible solutions for the double Ackermann mobility platform. Green arrows show the starting and goal locations, the green spline shows the resulting trajectory, and the dark blue arcs show path candidates for the expanded A* lattice nodes.}
    \label{fig:traj}
\end{figure}


The Kinematic Planner ingests \textit{transport triplets} as goal poses from the \textit{Transport Planner}. The goal poses from the \textit{transport triplets} are executed until completion, designated by a Euclidean distance and orientation threshold, at which point the next transport triplet is sent to the kinematic planner for continued operation.

The Kinematic Planner includes a lattice generator, an A* search algorithm \cite{Hart1968}, and tool and velocity planners. The lattice generator creates kinematically feasible path candidates, and A* is used to find a path within position and heading thresholds of a goal pose, as seen in Figure \ref{fig:traj}. Path candidates are assigned A* cost according to weighted topography cost and cumulative distance, with a heuristic of Euclidean distance between the terminal path position to the goal. The topography cost results in path solutions that minimize climbing heights to mitigate the risk of becoming high-centered during navigation. To improve online search efficiency, the heuristic was iteratively weighted to bias the search toward finding feasible solutions within online time constraints. If the robot deviates sufficiently from an existing plan, the kinematic planner replans a new trajectory from its current location to the existing goal pose. 

The velocity planner assigns a target velocity to each waypoint along the path, depending on whether the robot is traveling forward or backward. Finally, the tool planner assigns a target blade height to each waypoint using a heuristic further explained in Section \ref{sec:trajectory_control}. By integrating these planners, the kinematic planner can generate optimized and feasible trajectories for the robot to execute.


\subsection{Trajectory Control}\label{sec:trajectory_control}
The trajectory controller generates actuator commands for three objectives: steering, driving, and tool motion. Each of these objectives has its own specific controller. The steering controller, based on the Stanley control law \cite{hoffmann2007autonomous}, calculates the steer angle that can reduce and trade-off cross-track and heading error. The driving controller penalizes the driving velocity based on the steer speed and applies clamping and low-pass filters. Finally, the tool controller adjusts the tool position when the robot is either backing up or going forward to minimize undesired drag forces. The controller moves the tool down to a design height when going forward and up to an unobstructing height when backing up. This design height was set to graze flat ground when CraterGrader was placed onto the sand-filled testbed, described in Section \ref{sec:external-infra}.

\section{Details on Optimal Transport Planner}\label{sec:transport_details}


The Transport Planner solves the optimization problem of minimal-energy terrain manipulation to convert an arbitrary original topography to an arbitrary design topography. The final solution minimizes the total sum of volume moved multiplied by the distances traveled, analogous to mechanical work ignoring transport rate. The optimization is formulated as two LP problems that extend on typical optimal transport problems to account for non-equal volume distributions, as described in more detail in Section \ref{sec:comparison-to-standard}. Summarized in Equations (\ref{eq10})-(\ref{eq15}), the two linear programs differ only by their volume constraints, depending on two cases of (1) excess sink volume, or (2) excess source volume (or equal volumes). In our application, this approach manifests as a transport plan specifying how volume-surplus regions in the worksite (referred to as \textit{source nodes}) should be pushed to volume-lacking regions (referred to as \textit{sink nodes}), to ultimately produce flat, graded terrain. The planner is agnostic to input type, and in this work, nodes are formed from 2.5D height map grid cells. The sets of source and sink nodes form two distributions used in optimal transport. For grading of a cratered terrain, the primary source volume is crater rims and the primary sink volume is crater floors. The final transport plan results in pushing crater rims into the crater floors.  

 
 A simple example problem is provided in Appendix Section \ref{sec:example-transport-plan} with Figure \ref{fig:transport_toy_problem}. The solution to this sample problem yields an intuitive transport plan that assists with verification of the optimal transport algorithm.


\subsection{Key Assumptions}
The planner in this paper assumes that all volumes are homogeneous material; there is no accounting for specific material types. Future work could extend this formulation by accounting for material type as an additional cost parameter. The only costs factored into this planner are the volume scales and distance to move the volumes. Importantly, the planner also makes no assumptions about how volumes will be moved, so its use is not limited to the CraterGrader platform. The final transport plan could easily be used for machines with different terraforming modalities or distributed to multiple machines.


\subsection{Comparison to Standard Formulation}\label{sec:comparison-to-standard}
This planner extends on standard optimal transport problems by accounting for unequal distribution masses. Specifically, standard formulations either assume that distributions have the same integral areas, such as for probability distributions, or that the distributions are normalized to have the same total masses. However, inherent variability in terrain and inaccuracies in mapping mean that in practice there is nearly always unequal volume distributions. Normalizing these distributions would modify volume magnitudes which are required for scale-representative optimal transport. Maintaining these scales allows for insight into the estimated work required to execute the final plan. Allowing for unequal distribution sizes also naturally allows the planner to dictate where slack volume should remain; either ``high" source volumes left untouched when all ``low" sink volumes can be filled, or ``low" sink volumes that should remain unfilled because all ``high" source volumes are consumed. The practical application of this planner to real-world earth moving requires maintaining distribution scales and distances.



\subsection{Optimization Objective}
The optimization objective is to minimize the total volume transported between node pairs multiplied by the distances traveled. The volumes moved are continuous decision variables contained in a transport matrix $\Pi \in \mathbb{R}^{n \times m}$, where $n$ is the number of source nodes and $m$ is the number of sink nodes. $\Pi_{i,j}$ corresponds to the amount of volume that should be moved from the source node $i$ to the sink node $j$, in node volume units. In this work, node volume units were cubic meters. Distances between nodes are calculated prior to solving the optimization problem and are contained in a distance matrix $D \in \mathbb{R}^{n \times m}$, described further in Section \ref{sec:distance_matrix}. The objective then seeks to minimize the Frobenius inner product of $\Pi$ and $D$. This is provided in Equation (\ref{eq1}), with index notation used for clarity.

\begin{equation}
\min_{\Pi} \quad  \sum_{\substack{i=1,2,...,n\\j=1,2,...,m}} \Pi_{i,j} D_{i,j} \label{eq1}
\end{equation}

To simplify the formulation, an additional constraint enforces all transport volumes to be non-negative:

\begin{equation}
    -\Pi_{i,j} \leq 0 \quad \forall \quad i=1,...,n \text{ and } j=1,...,m\label{eq2}
\end{equation}

\subsection{Nodes and Volumes}
The set of $n$ source nodes is denoted $\bar{Y}$ and the set of $m$ sink nodes is denoted $\bar{X}$. Nodes are parameterized by planar worksite position coordinates $(p_x, p_y)$ and either source volume $v_y$ or sink volume $v_x$, where all volumes are defined as positive within their respective sets to simplify the optimization formulation. A source node $Y_i = \langle p_{x,i}, p_{y,i}, v_{y,i} \rangle$ is contained in the source set $\bar{Y} = \{Y_1, Y_2, ..., Y_n \}$, where $|\bar{Y}| = n$. Similarly, a sink node $X_j = \langle p_{x,j}, p_{y,j}, v_{x,j} \rangle $ is contained in the sink set $\bar{X} = \{X_1, X_2, ..., X_m \}$, where $|\bar{X}| = m$.

In the planner implementation, node volumes are estimated based on mapping data, where each grid cell in the map has a corresponding height. Each node volume is calculated using the grid area multiplied by the corresponding grid cell height. The grid cell area is a function of the map grid cell resolution but is a constant scalar for uniform cell sizes. The planner optimization formulation is based on volumes and assumes accounting for this calculation as data input. Note that with uniform cell sizes, a simplification could be made to use the height directly since all cells would be equally scaled by the same grid cell area.

\subsection{Distance Matrix}\label{sec:distance_matrix}
A distance matrix $D$, defined in Equation (\ref{eq3}), maintains the distances between planar positions of node pairs. Each element of the distance matrix $D_{i,j}$ corresponds to the Euclidean distance between source node $i$ and sink node $j$, and has the same units as the node coordinate units. In this work, the distance units were meters.

\begin{equation}
    D_{i,j} = d(Y_i, X_j) \quad \forall \quad i=1,...,n \text{ and } j=1,...,m \label{eq3}
\end{equation}

where

\begin{equation}
    d(Y_i, X_j)  = || {\begin{bmatrix}
        p_{x,i}\\
        p_{y,i}
    \end{bmatrix}} - {\begin{bmatrix}
        p_{x,j}\\
        p_{y,j}
    \end{bmatrix}}||_2 \label{eq4}
\end{equation}

\subsection{Volume Constraints}\label{sec:volume_constraints}
The volumes transported between nodes are constrained by the total magnitudes of source and sink volumes. In practice, these total volumes are likely to be unequal due to underlying site topography and mapping inaccuracies. To minimize mechanical work, we design constraints to allow for excess final volume from the larger of the source and sink volume magnitudes (i.e. unmoved source, or unfilled sink). Volume vectors are defined for the sets of source nodes $\vec{v}_y$ and sink nodes $\Vec{v}_x$ in Equation (\ref{eq5}).

\begin{equation}
\underset{n \times 1}{\vec{v}_y} = \begin{bmatrix}
        v_{y,1} \\
        v_{y,2} \\
        : \\
        v_{y,n}
    \end{bmatrix}
    ,\quad
   \underset{m \times 1}{\vec{v}_x} = \begin{bmatrix}
        v_{x,1} \\
        v_{x,2} \\
        : \\
        v_{x,n}
    \end{bmatrix} \label{eq5}
\end{equation}

Consider that the relative magnitudes of source and sink volumes can only be greater than, less than, or equal to each other. These conditions can be captured in two volume cases. Note that these cases could be captured in one unified optimization problem as a mixed integer linear program using a single binary decision variable, but for computational efficiency can instead be defined as two LP problems where only a single  LP needs to be solved online after checking the volume case.

Let Case 1 be defined as when sink volume is greater than source volume (i.e., $\sum_{i=1}^n v_{y,i} < \sum_{j=1}^m v_{x,j}$). The constraints should ensure that all sink volumes can remain unfilled but are not overfilled (inequality), while all source volumes are fully used (equality):

\begin{align}
    \Pi^\top \vec{\mathbf{1}}_n &\leq \vec{v}_x \label{eq6} \\
    \Pi \vec{\mathbf{1}}_m &= \vec{v}_y \label{eq7}
\end{align}

Similarly, let Case 2 be defined as when sink volume is the same or less than source volume (i.e., $\sum_{i=1}^n v_{y,i} \geq \sum_{j=1}^m v_{x,j}$). The constraints should then ensure that all source volumes can remain unused but are not overused (inequality), while all sink volumes are fully filled (equality):

\begin{align}
    \Pi \vec{\mathbf{1}}_m &\leq \vec{v}_y \label{eq8} \\
    \Pi^\top \vec{\mathbf{1}}_n &= \vec{v}_x \label{eq9}
\end{align}

\subsection{Final Optimization Problem}
Combining the objective with constraints gives the following final linear programming optimization problem formulation in Equations (\ref{eq10})-(\ref{eq15}). Note that this is a definition for two linear programs with identical objective (\ref{eq10}) and transport constraint (\ref{eq11}), where Case 1 uses additional constraints as Equations (\ref{eq12}) - (\ref{eq13}), and Case 2 uses additional constraints as Equations (\ref{eq14}) - (\ref{eq15}). The case constraints define the full respective LP, which can then be solved efficiently using any available LP solving methods. An example of solving this problem by vectorizing the decision variables is included in Appendix Section \ref{sec:transport-planner-cvxopt}.

\begin{alignat}{4}
    \min_{\Pi} &\quad&  \sum_{\substack{i=1,2,...,n\\ j=1,2,...,m}} \Pi_{i,j} D_{i,j} \label{eq10}\\
    \textrm{s.t.} & &  -\Pi_{i,j} \leq 0& \quad \forall i,j\label{eq11}\\
    \textit{Case 1 Constraints:} && \left( \sum_{i=1}^n v_{y,i} < \sum_{j=1}^m v_{x,j} \right) \nonumber \\
    & &\Pi^\top \vec{\mathbf{1}}_n \leq \vec{v}_x \label{eq12}\\
    & &\Pi \vec{\mathbf{1}}_m = \vec{v}_y \label{eq13}\\
    \textit{Case 2 Constraints:} && \left( \sum_{i=1}^n v_{y,i} \geq \sum_{j=1}^m v_{x,j} \right) \nonumber \\
    & &\Pi \vec{\mathbf{1}}_m \leq \vec{v}_y \label{eq14}\\
    & &\Pi^\top \vec{\mathbf{1}}_n = \vec{v}_x \label{eq15}
\end{alignat}
\\
where 
\begin{align*}
    \underset{n \times 1}{\vec{v}_y} = \begin{bmatrix}
        v_{y,1} \\
        v_{y,2} \\
        : \\
        v_{y,n}
    \end{bmatrix}, \quad
    \underset{m \times 1}{\vec{v}_x} = \begin{bmatrix}
        v_{x,1} \\
        v_{x,2} \\
        : \\
        v_{x,m}
    \end{bmatrix}\\
\end{align*}

\subsection{Runtime Analysis}


The overall problem size of the transport planner scales with the total number of nodes as $(n \times m)$, as computation complexity is dominated by optimizing for the continuous variables in the transport matrix $\Pi \in \mathbb{R}^{n\times m}$. Abstracted from solver specifics, the memory consumption scales with the problem size, $(n\times m)$. 

The actual online runtime performance depends primarily on (1) the problem size $(n \times m)$, (2) the distribution of node volumes (which determines solving difficulty), and (3) the optimization solver used. Full solutions for $\sim$1000 source and sink nodes ($\sim$2000 total) were observed to take approximately 30 seconds on a 2.5 GHz Quad-Core Intel Core i7 with 16 GB 1600 MHz DDR3 RAM using a C++ implementation with the GLOP solver from OR-Tools. An example of such a solution is included in Figure \ref{fig:chandrayaan_3_landing_site_crop_transport_plan}.

\subsection{Runtime Optimizations}
Implementation adjustments were made to improve online computational efficiency. Thresholding was applied to ignore source and sink nodes with volumes that were considered already close enough to the design topography. Additionally, in observing that the grading blade can push more than one cell at a time, a decimation filter was applied to the source nodes to enforce a minimum Euclidean distance between their planar coordinates. These two techniques effectively reduced the number of total nodes to speed up online computation. The threshold parameters were tuned for the anticipated crater size and grading blade width.


\begin{figure*}[]
    \medskip
    \centering
    \includegraphics[width=1.0\textwidth]{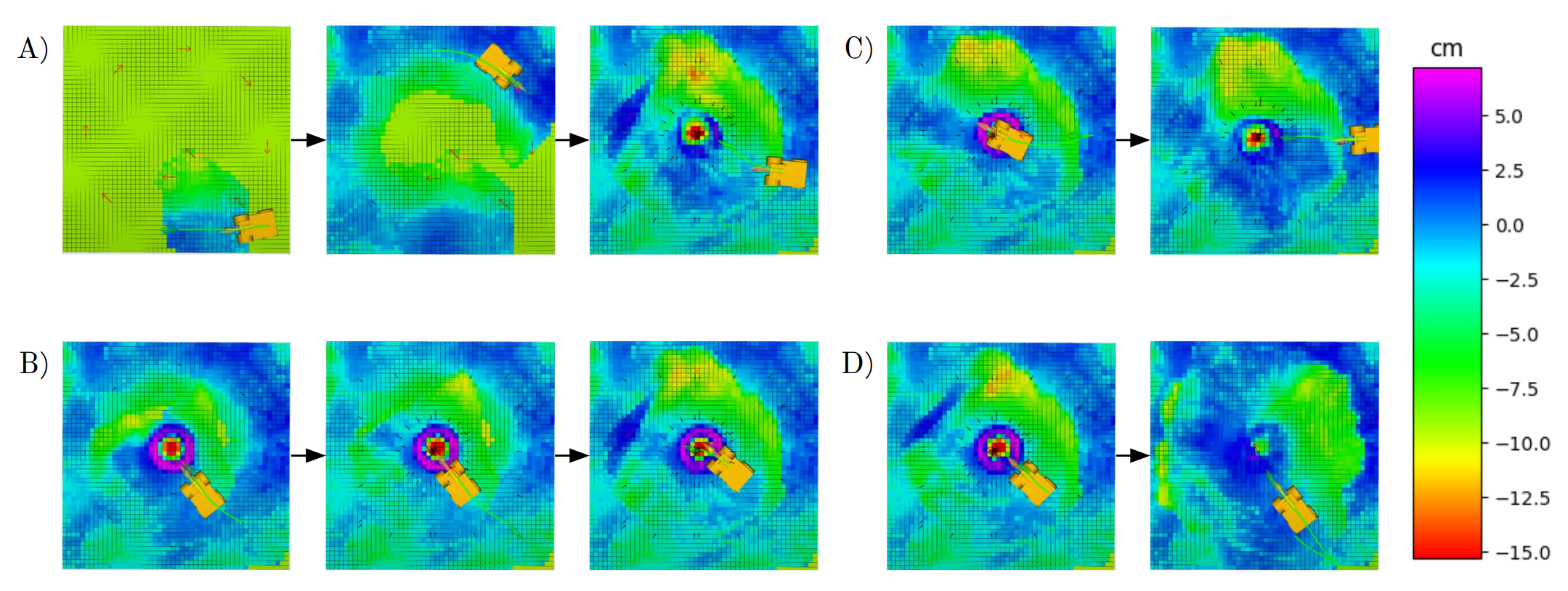}
    \caption{Autonomous site preparation with CraterGrader begins with an \textbf{A)} exploration pass of the worksite, guided by a set of waypoints (marked as red arrows) that span the worksite. \textbf{B)} Once the map is complete, the transport planner produces \textit{transport triplets}, (marked as black arrows) that act as waypoints dictating how the robot transports material from source nodes to sink nodes. \textbf{C)} CraterGrader reverses out of the crater to minimize difficult terrain traversal before continuing material transport. \textbf{D)} Illustrates all modifications to the site from beginning to end of operation to show how the crater feature has been predominantly cleared.}
    \label{fig:example_transport_plan}
\end{figure*}

\section{Experimental Results}

\subsection{Test Setup}
Testing was performed at the MoonYard described in Section \ref{sec:external-infra}, serving as an analog to lunar landscapes of interest for infrastructure development. Worksite terrain was initially prepared by manually smoothing to a grade within $\pm1\degree$, followed by digging a 1 m diameter crater near the center of the worksite, using a mold to create identical craters for repeatable testing. The crater diameter and depth-to-diameter ratio ($0.2$) were chosen to align within the LuSTR \cite{LUSTR} RFP specifications. Larger craters, in both depths and diameters, could require different sensor locations and perception approaches not explored in this work.

During evaluation, CraterGrader was placed into the prepared MoonYard. Fully autonomous mapping and grading operations, described in Section \ref{sec:system_objective}, were executed for an approximately 30-minute period, whereupon the robot was removed for worksite verification.

Verification was performed by capturing high-resolution point cloud scans before and after autonomous operation. A \textit{FARO Focus 3D} laser scanner generated scans of worksite terrain accurate to millimeter precision, which were processed offline in an open-source 3D data analysis software. Out-of-worksite data points were removed from the point cloud and a ground plane was fitted from the remaining points. The data was then exported to an analysis pipeline to compute validation metrics such as surface grade, smoothness, and area OOS reduction. 

\subsection{Results}

System performance was evaluated across five 30-minute trials of end-to-end autonomous operation, including a live demonstration. Average performance results for the four evaluation trials are shown in Table \ref{tab:avg-performance}, and results for the live performance are depicted in Table \ref{tab:live-performance}. The key metrics to measure performance are selected as worksite grade, worksite smoothness, and area out-of-spec (OOS) reduction.

Grade and smoothness specifications are derived from the NASA LuSTR RFP \cite{LUSTR}, which define grade and smoothness requirements as $\pm1\degree$ from a fit plane and 1 cm in height standard deviation respectively. Figure \ref{fig:crater-profile} illustrates the definitions of grade angle and smoothness with respect to a crater profile. Grade is defined as the angle between a plane fit to the terrain profile, and a level plane normal to gravity. Smoothness is calculated from the standard deviation of a distribution of terrain heights relative to the fit plane. On average, CraterGrader achieved a worksite grade of 0.11$\degree$ and smoothness of 0.7 cm. In a culminating live demonstration for a viewing audience, CraterGrader achieved 0.07$\degree$ grade and 0.6 cm smoothness. In all cases, CraterGrader is able to repeatably produce a smoothly graded worksite that meets the desired NASA baseline requirements in \cite{LUSTR}.

\begin{figure}[]
    \medskip
    \centering
    \includegraphics[width=0.45\textwidth]{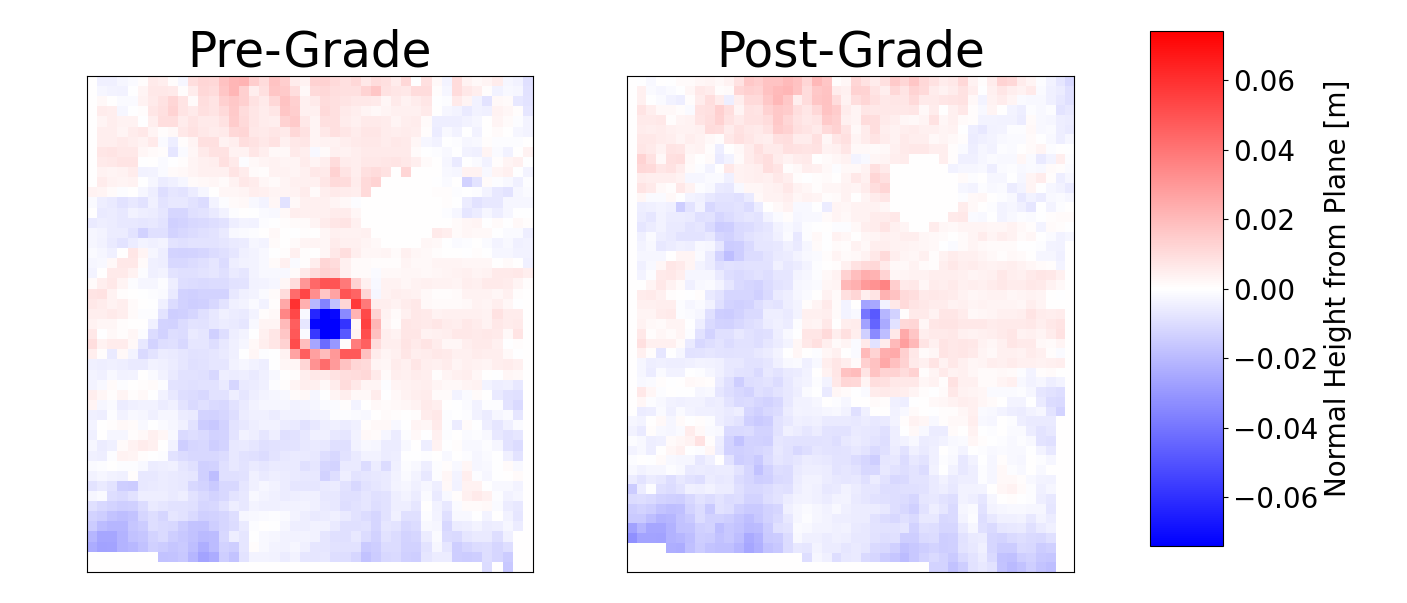}
    \caption{Pre-Grade vs. Post-Grade verification map of a 5 m x 5 m worksite with a single 1 m diameter crater, comparing worksite topography before and after a 30-minute fully autonomous grading operation.}
    \label{fig:vav}
\end{figure}

The area OOS reduction metric measures the percent reduction in worksite area that lies outside of the desired $\pm1\degree$ grade and 1 cm smoothness requirements before and after autonomous operation. Although there are no specific requirements for area OOS reduction per the NASA LuSTR RFP \cite{LUSTR}, this metric captures the essence of CraterGrader's ability to manipulate unstructured terrain, and transform it into a worksite that meets a set of desired specifications. On average, CraterGrader achieved an 88\% reduction in area OOS (from 0.45 m$^2$ to 0.05 m$^2$) for all four evaluation trials. CraterGrader also achieved an 82\% reduction in area OOS (from 0.45 m$^2$ to 0.08 m$^2$) during the live demonstration, measured from the average starting conditions of the prior evaluation trials. Figure \ref{fig:vav} illustrates an example worksite terrain topography from the evaluation trials, both pre- and post-grade, in which the prominent crater feature was flattened after autonomous operation.

\begin{table}[]
    \centering
    \medskip
    \begin{tabular}{|c|c|c|c|}
    \hline
    \textbf{Metric} & \textbf{Desired} & \textbf{Pre-Grade} & \textbf{Post-Grade} \\ \hline
    Worksite Grade & $\pm1\degree$ & 0.12\degree & 0.11\degree \\ \hline
    Worksite Smoothness & $<$ 1 cm & 1.25 cm & 0.7 cm \\ \hline
    Area OOS Reduction & N/A & N/A & 88\% \\ \hline
    \end{tabular}
    \caption{CraterGrader Average Autonomous Grading Performance}
    \label{tab:avg-performance}
\end{table}

\begin{table}[]
    \centering
    \begin{tabular}{|c|c|c|}
    \hline
    \textbf{Metric} & \textbf{Desired} & \textbf{Post-Grade} \\ \hline
    Worksite Grade & $\pm1\degree$ & 0.07\degree \\ \hline
    Worksite Smoothness & $<$ 1 cm & 0.6 cm \\ \hline
    Area OOS Reduction & N/A & 82\% \\ \hline
    \end{tabular}
    \caption{CraterGrader Live Autonomous Grading Performance}
    \label{tab:live-performance}
\end{table}

\subsection{Additional Simulation}

\begin{figure}[]
    \medskip
    \centering
    \includegraphics[width=0.45\textwidth]{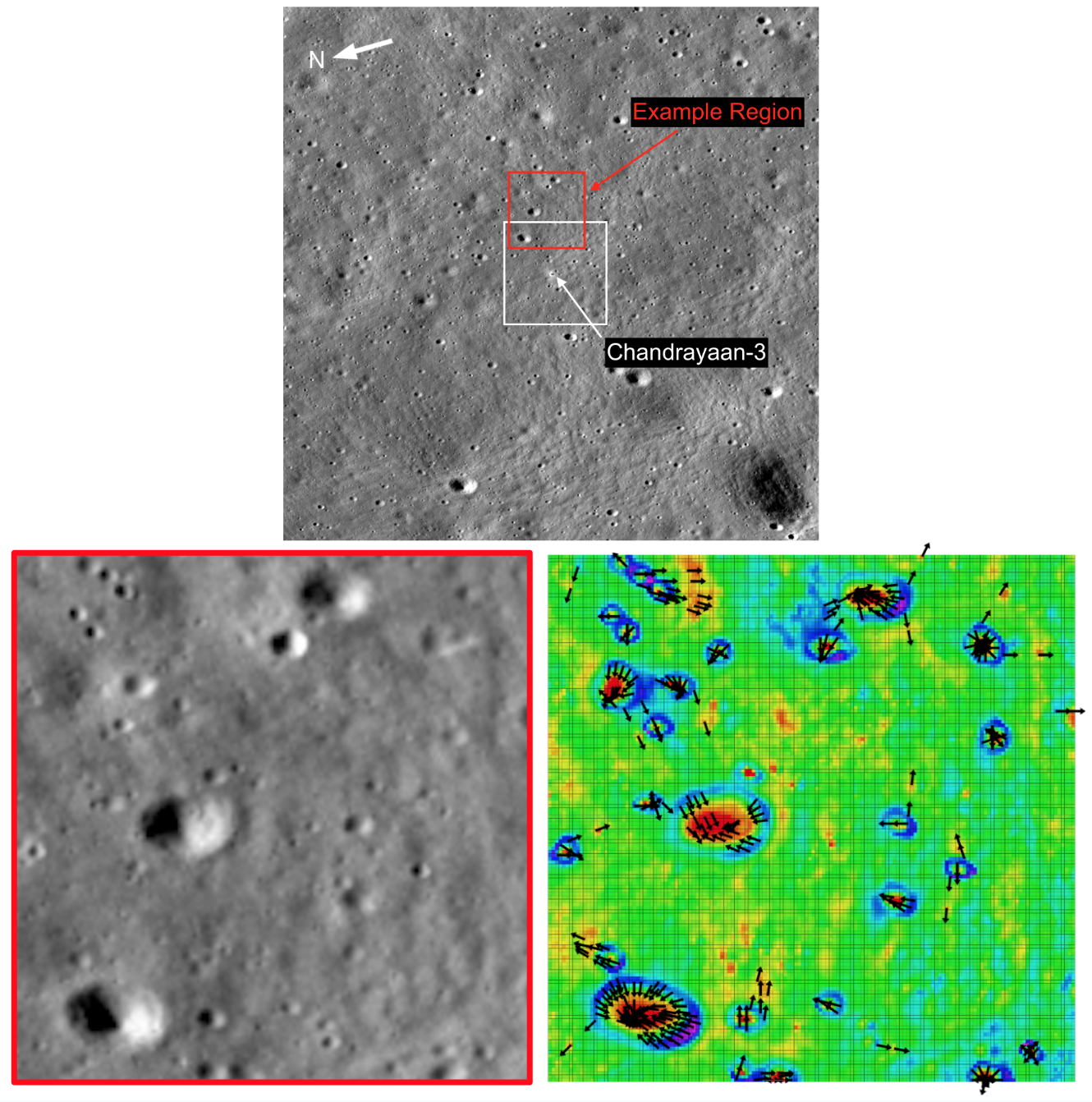}
    \caption{Transport plan applied to a section of the Chandrayaan-3 Landing Site captured by Lunar Reconnaissance Orbiter Camera \cite{asuChandrayaan3Landing}. The black arrows on the right image indicate material transport.}
    \label{fig:chandrayaan_3_landing_site_crop_transport_plan}
\end{figure}

In addition to hardware tests, additional simulations were run to demonstrate transport planner effectiveness for larger maps with more direct applicability to real world scenarios. An example simulation depicting the Chandrayaan 3 landing site \cite{asuChandrayaan3Landing} is included in Figure \ref{fig:chandrayaan_3_landing_site_crop_transport_plan}. The worksite region used in this simulation is roughly 300 meters square. In lieu of absolute height scales (which are uncertain due to the lack of available high-resolution lunar height maps), terrain height is modeled using image pixel intensity, for the purposes of demonstrating transport planner functionality.

The simulation is designed to challenge the transport planner with a more complex terrain topography, better representative of a hypothetical lunar worksite. This expands upon the simpler terrain profiles used in hardware demonstrations. The resulting transport plan illustrates appropriate material transport from regions of high terrain to low terrain, pushing crater rims into crater floors. The simulation also shows how the transport plan seamlessly handles arbitrary terrain formations such as ridges and non-uniform craters. 

\section{Lessons Learned \& Key Takeaways}

\subsection{Patient work allows for iterative improvement with simplified hardware}

\begin{itemize}
    \item  Since lunar robotics applications have a lessened focus on time efficiency compared to robustness, accuracy, and energy efficiency, it is valuable to focus on iterative operations at smaller and simpler scales. Slower, but simplified approaches can result in a more robust solution. Even in a mass-constrained system, the shallow cuts of CraterGrader's grading blade provide exceptional results due to the allowance of multiple passes and self-corrections.
\end{itemize}

\subsection{External infrastructure provides key localization accuracy in dynamic, near-featureless environments}

\begin{itemize}
    \item There is significant value in providing external infrastructure to provide the required precision for site preparation with autonomous graders. The use of proprioceptive localization alone is limiting, especially in large environments with global accuracy requirements, due to similar lunar features, compounding error inherent in relative state sensors, and the inherent transience of site features during terrain modification. Accurate and high-precision position estimation in the z-direction is of particular importance to this application, to enable precise mapping of terrain heights and tool positioning during earthmoving operations. The robotic total station, fixed external to the worksite, resulted in highly accurate 3D positioning, independent of the feature richness in the surrounding environment.
\end{itemize}

\subsection{Planning with a generic representation of the environment leads to flexible application and ease of error correction}
\begin{itemize}
    \item Representing the worksite as a 2.5D height map which is ingested by the high-level task planner allows for a wide variety of site states to be acted upon. This generalization is valuable for unstructured terrain modification. Additionally, CraterGrader is able to self-correct after analyzing errors in the current state of the worksite with minimal additional behavior paths. Designing the planner with explicit primitives, such as detecting craters in the worksite, would have led to more challenging recovery behaviors.
\end{itemize}

\section{Conclusions and Future Work}

CraterGrader is a novel approach for modifying unstructured terrain to achieve a desired topography with planetary robotics constraints in mind. Unstructured and continuously deforming terrain is perceived and dynamically mapped, leveraging precision GPS-free localization subject to lunar constraints. Tool and motion planning is performed by calculating a transport plan online using a linear programming formulation of earth mover's distance and then executed with tool, drive, and steer control.

This work aims to inform future planetary site preparation autonomy development, in addition to any terrestrial applications which may benefit from similar levels of autonomy. The demonstration of our low-mass rover has shown that this approach is feasible, robust to uncertain dynamics and measurements, and well-suited to autonomous site preparation and earthmoving for a lunar environment. This represents a significant step toward the future of autonomous robots for lunar site preparation and intelligent, energy-efficient terrain manipulation in general. Future work could include introducing additional modalities of earthmoving tasks such as rock picking, compaction, drilling, blasting, etc. to create a more versatile machine. It is also in the interest of the authors to extend the work of the machine-agnostic transport planner to other heterogenous robotic vehicles for material transport which could take the form of a scraper, bulldozer, backhoe, excavator, loader, or dump truck, possibly in multi-agent applications.

\section*{ACKNOWLEDGMENT}

We would like to give special thanks to Dr. John Dolan, Warren ``Chuck" Whittaker, Dr. Dimitrios Apostolopoulos, Dr. David Wettergreen, and the MRSD (Master of Science in Robotic Systems Development) program of the Carnegie Mellon University Robotics Institute for their support, opinions, and advice throughout this journey. 

\appendix


\subsection{Additional Media}\label{sec:additional-media}

A recording of a test run of the CraterGrater worksystem can be seen at \href{https://www.youtube.com/watch?v=8xV1q1SpIpc}{youtube.com/watch?v=8xV1q1SpIpc}.

\subsection{Example Transport Planner Problem}\label{sec:example-transport-plan}

Consider an example problem with source set $\bar{Y}$ and sink set $\bar{X}$ defined by:

\begin{align*}
    \bar{Y} = \{& \langle p_{x,1}=-1, p_{y,1}=-0.5, v_{y,1}=0.2 \rangle,\\
    &\langle p_{x,2}=0.5, p_{y,2}=-1, v_{y,2}=0.6 \rangle\}\\
    \bar{X} = \{& \langle p_{x,1}=-2, p_{y,1}=1, v_{x,1}=0.3 \rangle,\\
    &\langle p_{x,2}=2, p_{y,2}=1, v_{x,2}=0.4 \rangle\}
\end{align*}

\begin{figure}[h!]
    \medskip
    \centering
    \includegraphics[width=0.45\textwidth]{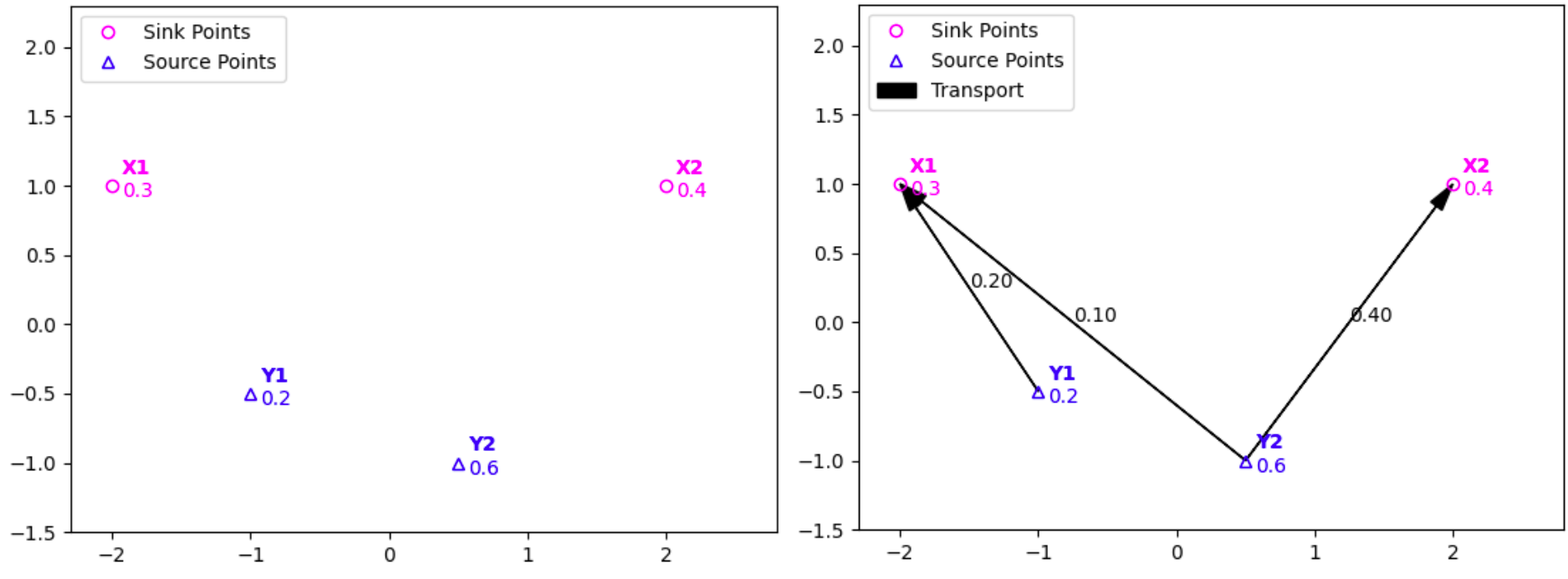}
    \caption{Simple example problem for transport planner.}
    \label{fig:transport_toy_problem}
\end{figure}

In this case, there is more source volume than sink volume, as the total source volume sums to 0.8 and the total sink volume sums to 0.7. A visualization of the solution is in Figure \ref{fig:transport_toy_problem}. The optimal solution transports 0.2 volume from $Y_1$ to $X_1$, 0.1 volume from $Y_2$ to $X_1$, 0.4 volume from $Y_1$ to $X_2$, and leaves the excess 0.1 source volume at node $Y_1$. This solution minimizes the total volume moved and aligns with intuition.

\subsection{Transport Planner Linear Program Formulation}\label{sec:transport-planner-cvxopt}

One way in which the transport planner optimization problem can be solved using linear programming is by converting the objective and constraints to block matrices and vectors for standard form.

\begin{alignat*}{2}
    &\text{minimize} \quad\quad && c^T \hat{x}\\
    &\text{subject to}          && G\hat{x} \leq h \\ 
    &                           && A\hat{x} = b
\end{alignat*}

To convert the objective, we need to turn the sum $\sum_{\substack{i=1,2,...,n\\ j=1,2,...,m}} \Pi_{i,j} D_{i,j}$ into a dot product between a vector $c$ of constant cost terms and a vector $\hat{x}$ of the optimization variables. The distance matrix can be converted into the constant cost vector and the transport matrix can be converted into the continuous optimization variable vector using row-major order as follows:

\begin{equation}
    \underset{nm \times 1}{c} = \begin{bmatrix}
        d(Y_1, X_1)\\
        d(Y_1, X_2)\\
        :\\
        d(Y_2, X_1)\\
        d(Y_2, X_2)\\
        :\\
        d(Y_n, X_m)
    \end{bmatrix}, \quad
    \underset{nm \times 1}{\hat{x}} = \begin{bmatrix}
        \Pi_{1,1}\\
        \Pi_{1,2}\\
        :\\
        \Pi_{2,1}\\
        \Pi_{2,2}\\
        :
        \Pi_{m,n}\\
    \end{bmatrix}
\end{equation}

To convert the constraints, we can define block matrices for creating the matrices $G, A$ and vectors $h, b$ for inequality and equality constraints, respectively. Note that the following formulations again assume row-major order.

To replicate the $\Pi \vec{\mathbf{1}}_m$ term, we can make a one's block matrix $B_{1,m}$ packed with the all-ones vector $\vec{\mathbf{1}}_m^\top$ along the block diagonal. There need to be $n$ block columns in total to have zeros applied to the rest of the optimization variable vector $\hat{x}$, and $n$ block rows to sum for all $n$ source nodes, making the final dimensions $B_{1,m} \in \mathbb{R}^{n \times (m \times n)}$. 

\begin{equation}
    \underset{n \times mn}{B_{1,m}} = \begin{bmatrix}
        \vec{\mathbf{1}}_m^\top & 0 & ... & 0 \\
        0 & \vec{\mathbf{1}}_m^\top & ... & 0 \\
        : & ... & \ddots & 0 \\
        : & ... & ... & \vec{\mathbf{1}}_m^\top \\
    \end{bmatrix}
\end{equation}

Similarly, to replicate the $\Pi^\top \vec{\mathbf{1}}_n$ term in the matrix $G$, we can make a one's block matrix $B_{1,n}$ packed with $n$ identity matrices. This then applies zeros to all of the optimization variable vector $\hat{x}$ except for what would be each column in the transport matrix. 

\begin{equation}
    \underset{m \times mn}{B_{1,n}} = \begin{bmatrix}
        \underset{m \times m}{I} & \underset{m \times m}{I} & ... & \underset{m \times m}{I}\\
    \end{bmatrix}
\end{equation}

These block matrices $B_{1,m}$ and $B_{1,n}$ can be used for the constraints \ref{eq12}, \ref{eq13}, \ref{eq14}, \ref{eq15}.

The last remaining constraint \ref{eq11} is then to enforce that all transport volumes are positive or zero, which can be done using an $(nm \times nm)$ identity matrix $I_{\Pi}$ with zero's vector $\mathbf{0}$ of length $nm$:

\begin{equation}
    \underset{nm \times nm}{I_{\Pi}} {\hat{x}} \geq \underset{nm \times 1}{\mathbf{0}}
\end{equation}

The final LP optimization problem can then be summarized as the following:

\begin{alignat}{2}
    &\text{minimize} \quad\quad && c^T \hat{x}\\
    &\text{subject to} && G\hat{x} \leq h \\
    &                  && A\hat{x} = b
\end{alignat}

where for Case 1 when there is more sink than source $\left( \sum_{i=1}^n v_{y,i} < \sum_{j=1}^m v_{x,j} \right)$, we have:

\begin{align}
    \underset{(nm+m) \times nm}{G} &= \begin{bmatrix}
       -I_{\Pi} \\
        B_{1,n}
    \end{bmatrix}
\\
    \underset{(nm+m) \times 1}{h} &= \begin{bmatrix}
        \mathbf{0} \\
        \vec{v}_x 
    \end{bmatrix}
\\
    \underset{n \times nm}{A} &= \begin{bmatrix}
        B_{1,m} 
    \end{bmatrix}
\\
    \underset{n \times 1}{b} &= \begin{bmatrix}
        \vec{v}_y 
    \end{bmatrix}
\end{align}
while for Case 2 when there is more source than sink, or equal amounts, $\left( \sum_{i=1}^n v_{y,i} \geq \sum_{j=1}^m v_{x,j} \right)$, we have:
\begin{align}
    \underset{(nm+n) \times nm}{G} &= \begin{bmatrix}
       -I_{\Pi} \\
        B_{1,m}
    \end{bmatrix}
\\
    \underset{(nm+n) \times 1}{h} &= \begin{bmatrix}
        \mathbf{0} \\
        \vec{v}_y 
    \end{bmatrix}
\\
    \underset{m \times nm}{A} &= \begin{bmatrix}
        B_{1,n} 
    \end{bmatrix}
\\
    \underset{m \times 1}{b} &= \begin{bmatrix}
        \vec{v}_x 
    \end{bmatrix}
\end{align}

The final transport solution is then contained in the optimization variable vector $\hat{x}$, which can be converted back to the full transport matrix $\Pi$ by reshaping it to $n\times m$ using row-major order.

\bibliography{references}

\end{document}